%% file: PaperForReview.tex
\documentclass[10pt,twocolumn,letterpaper]{article}

\usepackage{wacv}              %
\usepackage[accsupp]{axessibility}  %

\usepackage[T1]{fontenc}
\usepackage[utf8]{inputenc}
\usepackage{graphicx}
\usepackage{amsmath}
\usepackage{amssymb}
\usepackage{booktabs}
\usepackage{enumitem}
\usepackage{multirow}
\usepackage[normalem]{ulem}
\useunder{\uline}{\ul}{}
\usepackage[pagebackref,breaklinks,colorlinks]{hyperref}
\usepackage{flushend}

\usepackage[capitalize]{cleveref}
\crefname{section}{Sec.}{Secs.}
\Crefname{section}{Section}{Sections}
\Crefname{table}{Table}{Tables}
\crefname{table}{Tab.}{Tabs.}

\usepackage[accsupp]{axessibility}  %

\def\wacvPaperID{1880} %
\def\confName{WACV}
\def\confYear{2024}

\AddToShipoutPicture*{%
     \AtTextUpperLeft{%
         \put(-3.5,10){
           \begin{minipage}{\textwidth}
              \scriptsize
              \MakeUppercase{Preprint version of an IEEE/CVF Winter Conference on Applications of Computer Vision (WACV) paper; final version available at:} \url{https://ieeexplore.ieee.org}
           \end{minipage}}%
     }%
}

\begin{document}

\title{FocusTune: Tuning Visual Localization through Focus-Guided Sampling}

\author{Son Tung Nguyen \qquad Alejandro Fontan \qquad Michael Milford \qquad Tobias Fischer\\[0.25cm]
Queensland University of Technology\\
Brisbane, Australia\\
{\tt\small sontung.nguyen@hdr.qut.edu.au}
}

\maketitle

\newcommand{\est}[1]{\hat{#1}}
\newcommand{\gt}[1]{{#1}^*}

\newcommand{\pos}{\mathbf{y}}
\newcommand{\eye}{\mathbf{e}}
\newcommand{\pixel}{\mathbf{y}}
\newcommand{\patch}{\mathbf{p}}
\newcommand{\desc}{\mathbf{d}}

\newcommand{\crd}{\mathbf{x}}
\newcommand{\crds}{\mathcal{X}}

\newcommand{\pixels}{\mathcal{Y}}
\newcommand{\instance}{\mathcal{G}}

\newcommand{\image}{\mathit{I}}
\newcommand{\buffer}{\mathbf{B}}
\newcommand{\radius}{\rho}

\newcommand{\extrinsic}{H}
\newcommand{\intrinsic}{K}

\begin{abstract}
We propose FocusTune, a focus-guided sampling technique to improve the performance of visual localization algorithms. FocusTune directs a scene coordinate regression model towards regions critical for 3D point triangulation by exploiting key geometric constraints. Specifically, rather than uniformly sampling points across the image for training the scene coordinate regression model, we instead re-project 3D scene coordinates onto the 2D image plane and sample within a local neighborhood of the re-projected points. While our proposed sampling strategy is generally applicable, we showcase FocusTune by integrating it with the recently introduced Accelerated Coordinate Encoding (ACE) model. Our results demonstrate that FocusTune both improves or matches state-of-the-art performance whilst keeping ACE's appealing low storage and compute requirements, for example reducing translation error from 25 to 19 and 17 to 15 cm for single and ensemble models, respectively, on the Cambridge Landmarks dataset. This combination of high performance and low compute and storage requirements is particularly promising for applications in areas like mobile robotics and augmented reality. We made our code available at \url{https://github.com/sontung/focus-tune}.
\end{abstract}

\input{1-introduction}
\input{2-related_works}
\input{3-problem_statement}
\input{4-prelimilary}
\input{5-methodology}

\input{6-results}
\input{7-conclusion}
\section*{Acknowledgements}This research was partially supported by funding from ARC Laureate Fellowship FL210100156 to MM, the QUT Centre for Robotics, and Intel Research via grant RV3.290.Fischer.

{\small
\bibliographystyle{ieee_fullname}
\bibliography{egbib}
}

\end{document}

%% file: 1-introduction.tex
\vspace{-0.3cm}
\section{Introduction}
\label{sec:intro}

\begin{figure*}[ht]
    \centering
    \includegraphics[width=1\linewidth]{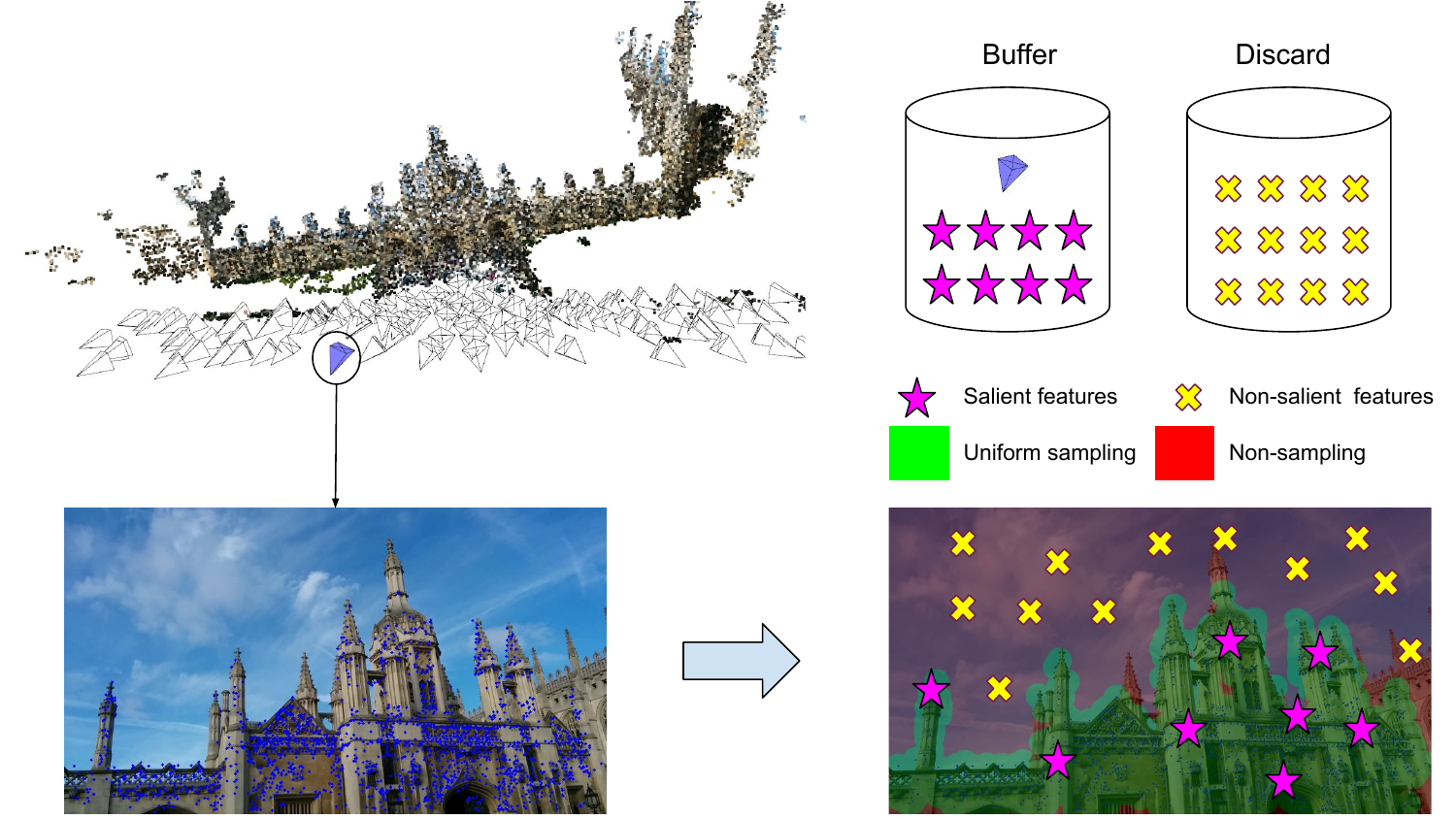}
    \caption{\textbf{System Overview.} For every training image, using the associated camera pose \textit{(purple color)} we re-project the map \textit{(top-left image)} onto that image to obtain a set of seed keypoints \textit{(blue dots in the bottom-left image)}. We then sample training instances uniformly within a region of $\radius$ pixels surrounding the seed keypoints \textit{(green region of the bottom-right image)} and prohibit sampling from outside these circles \textit{(red region of the bottom-right image)}. As a result, our training instances are sampled from important regions of the image \textit{(pink stars in the bottom-right image)} rather than texture-less regions \textit{(yellow crosses in the bottom-right image)}.}
    \label{fig:ps_rq1:system_overview}
    \vspace*{-0.35cm}
\end{figure*}

Visual localization remains a cornerstone task in computer vision, with substantial ramifications for applications in mobile robotics~\cite{heng2019project, lim2012real} and augmented reality (AR)~\cite{arth2011real, castle2008video}. The primary objective is to determine the pose -- both position and orientation -- of a query image relative to a pre-established environment map. Such maps typically comprise three components: a set of RGB images, a 3D point cloud approximating the environment's geometry, and individual camera poses corresponding to each image.

Historically, structured methods have been employed to solve the visual localization problem, utilizing feature correspondences between 2D pixels in the query image and 3D points within the environmental map~\cite{DBLP:conf/eccv/LiSH10, active_search, sattler2015hyperpoints}. To manage the computational complexity, these methods often resort to clustering and quantizing the search space. While they offer excellent accuracy and low query times, their memory footprint can be prohibitive for large-scale environments. In this paper, we further close the performance gap to structured methods while retaining storage and training efficiency.

Modern regression-based techniques offer an alternative to structured methods by embedding the environmental map into the neural network's architecture, thus sidestepping the memory constraint. Although Absolute Pose Regression (APR) models~\cite{DBLP:conf/iccv/KendallGC15, DBLP:conf/iccv/ShavitFK21, DBLP:journals/corr/WalchHLSHC16, li2gtcar, zhou2022geometry}, which directly regress the camera pose, have gained traction, they suffer from accuracy limitations, mainly due to unresolved questions around their optimal training objectives~\cite{DBLP:journals/pami/BrachmannR22, kendall2017geometric, DBLP:conf/iccv/KendallGC15, brahmbhatt2018geometry}.

Scene Coordinate Regression (SCR) models~\cite{DBLP:conf/cvpr/ShottonGZICF13, brachmann2023accelerated, DBLP:journals/pami/BrachmannR22, brachmann2016uncertainty, brachmann2019expert, DBLP:conf/3dim/CavallariBMTG19, do2022learning, huang2021vs, li2020hierarchical} represent a marked advancement in this regard, trading pose regression for scene coordinate regression and achieving higher accuracy. These models regress the 3D coordinates of the 2D pixels in an image, as opposed to directly regressing the camera pose as in APR models, which allows them to be trained via the re-projection error.

One high-performing popular SCR model is DSAC*~\cite{DBLP:journals/pami/BrachmannR22}, where SCR was combined with differentiable RANSAC. However, the training time for DSAC*~\cite{DBLP:journals/pami/BrachmannR22} can be up to 15 hours for a given scene. Recently, the Accelerated Coordinate Encoding (ACE) model~\cite{brachmann2023accelerated} emerged as a rapid training alternative, substantially reducing computational time to just 5 minutes by splitting the network into two parts: a scene-agnostic pre-trained backbone and scene-specific regression heads. ACE relies on a training buffer from which high-dimensional feature vectors are extracted for various regions in an image.

In this paper, we demonstrate that ACE's training buffer contains a considerable number of non-informative feature vectors, thus hindering the optimal training of the scene-specific regression heads. Our proposed FocusTune heuristic sampler guides the training process to focus on important parts of the scene, specifically those that form the 3D point cloud map. This results in marked performance improvements with minimal overhead to the training time. The sampling process is illustrated in Fig.~\ref{fig:ps_rq1:system_overview}.

\noindent We summarize our contributions as follows:
\begin{enumerate}[topsep=0pt,itemsep=-1ex,partopsep=1ex,parsep=1ex]
    \item We introduce FocusTune, a focus-guided heuristic sampling strategy to avoid optimizing parts of the scene that are non-informative and instead leveraging geometrical constraints. FocusTune incurs minimal computational overhead.
    \item We validate and implement FocusTune within the ACE framework~\cite{brachmann2023accelerated}, thereby exemplifying FocusTune's compatibility and effectiveness in a real-world setting.
    \item We provide empirical evidence showing marked performance gains, including a reduction in translation error from 25cm to 19cm on the Cambridge Landmarks dataset when using a single model, approaching the performance of computationally intensive methods like DSAC*. When using an ensemble scheme, FocusTune results in a new state-of-the-art translation error of just 9cm across all prior visual localization techniques, including structured methods that require substantially more storage.
\end{enumerate}

%% file: 2-related_works.tex
\section{Related Works}
\subsection{Feature Detection}
Feature detection serves as a foundational element in numerous geometric computer vision applications, such as Simultaneous Localization and Mapping (SLAM)\cite{cadena2016past} and Structure from Motion (SfM)\cite{schoenberger2016sfm}. These applications rely on the robust identification of pixels, or features, that exhibit consistent characteristics across varying lighting conditions and viewpoints. Such features typically originate from 2D pixels with distinct textural properties, as opposed to texture-deficient uniform areas. The conventional approach for feature detection employs the Difference-of-Gaussian technique~\cite{DBLP:journals/ijcv/Lowe04}. This method involves iteratively smoothing the input image across multiple scales and computing the differences between successive scales. Pixels that remain sharp through this process are designated as features. Owing to its computational efficiency, this technique has been widely adopted in prevalent feature description algorithms~\cite{DBLP:journals/ijcv/Lowe04, DBLP:conf/eccv/BayTG06, mikolajczyk2004scale}.

More recent developments~\cite{detone2018superpoint, noh2017large, dusmanu2019d2} have explored the use of neural networks for feature detection. For example, the method described in \cite{noh2017large} utilizes an attention layer to aggregate the semantic significance of local features, thereby facilitating the identification of the most relevant features while also providing reliable confidence scores. Conversely, \cite{detone2018superpoint} presents a learning-based corner detection model trained on a large synthetic dataset comprising basic shapes like triangles and cubes.

Our work builds upon this expansive body of literature, demonstrating that learning-based localization systems can also achieve enhanced performance by focusing on textured regions during training.

\subsection{Structured Methods}
Classical structured solutions~\cite{DBLP:conf/cvpr/IrscharaZFB09, DBLP:conf/eccv/LiSH10, active_search, sattler2015hyperpoints, sattler2016efficient} for visual localization tasks establish 2D-3D correspondences between query images and environment maps. The nearest-neighbor search is typically done by comparing the feature descriptors~\cite{DBLP:journals/ijcv/Lowe04} of the query image and the descriptors of the 3D points. Therefore, structured methods require access to the descriptor database of the map at run time, making them unsuitable for memory-constrained devices. While GoMatch~\cite{zhou2022geometry} eliminates the need for storing descriptors, this comes at the expense of reduced accuracy.

\subsection{Absolute Pose Regression}
The advent of deep learning has enabled absolute pose regression (APR) models to directly regress camera poses from query images~\cite{DBLP:conf/iccv/KendallGC15, DBLP:conf/iccv/ShavitFK21, DBLP:journals/corr/WalchHLSHC16, li2gtcar, zhou2022geometry}. APR models output the rotation (e.g.~represented by a quaternion) and the translation vectors of the corresponding camera pose to the input image. While earlier methods \cite{DBLP:conf/iccv/KendallGC15} relied on a pre-trained GoogLeNet~\cite{DBLP:conf/cvpr/SzegedyLJSRAEVR15} as the backbone, more recent network architectures increase the performance of APR models with spatial LSTMs~\cite{DBLP:journals/corr/WalchHLSHC16} or transformers~\cite{DBLP:conf/iccv/ShavitFK21, li2gtcar, tang2023neumap}. However, despite being fast and requiring no access to maps at query time, the accuracy of APR models is still far behind classical structured methods.

\subsection{Scene Coordinate Regression}
Scene coordinate regression models predict the scene coordinates for the pixels in the query images~\cite{DBLP:conf/cvpr/ShottonGZICF13, DBLP:conf/cvpr/ValentinNSFIT15, DBLP:journals/pami/BrachmannR22, cavallari2017fly, li2020hierarchical}. SCR models can be implemented by random forests~\cite{DBLP:conf/cvpr/ShottonGZICF13, DBLP:conf/cvpr/ValentinNSFIT15} or convolutional neural networks~\cite{DBLP:journals/pami/BrachmannR22, brachmann2023accelerated, brachmann2016uncertainty, brachmann2018learning, brachmann2019expert, do2022learning, huang2021vs}. SCR models are often optimized via the re-projection error, which works well in practice, thus leading to a much higher accuracy than APR models that are competitive with structured methods in certain settings. Up until recently, the mapping time of SCR models was problematic, often requiring hours of mapping time~\cite{DBLP:journals/pami/BrachmannR22, brachmann2019expert} for convergence.

\subsection{Low Mapping Time Scene Coordinate Regression}
Multiple approaches were proposed in the literature to reduce the mapping time of SCR models~\cite{brachmann2023accelerated, dong2022visual, DBLP:conf/cvpr/ShottonGZICF13, cavallari2019real, yang2019sanet}. SCoRF~\cite{DBLP:conf/cvpr/ShottonGZICF13} showed a random forest can localize well in just 10 minutes. \cite{cavallari2019real} leverages depth information to further increase the accuracy. SAnet~\cite{yang2019sanet} requires access to database scene images to retrieve and interpolate database scene coordinates. Despite being scene-agnostic and having a short mapping time, its weaknesses include higher memory requirement and lower accuracy compared to classical structured methods. 

\cite{brachmann2023accelerated, dong2022visual} recently proposed to divide the SCR network into two parts: scene-agnostic pre-trained backbone and scene-specific regression heads. The backbone is trained once on a large dataset~\cite{dai2017scannet}, whereas the regression heads are trained specifically for a given scene. \cite{dong2022visual} uses another convolutional layer to implement the regression head. This design choice cuts the mapping time to only 2 minutes, however, increases the memory footprint and decreases the accuracy. On the other hand, ACE~\cite{brachmann2023accelerated} uses multi-layer perceptron layers to realize the regression heads, leading to state-of-the-art results on both indoor and outdoor settings with only RGB mapping images.

In this paper, we show that it is possible to further extend the accuracy of ACE models with a small overhead at training time by heuristic sampling of the training buffer.

%% file: 3-problem_statement.tex
\section{Preliminaries}
\subsection{Problem Statement}

\begin{figure*}[t]
    \centering
    \includegraphics[width=1\linewidth]{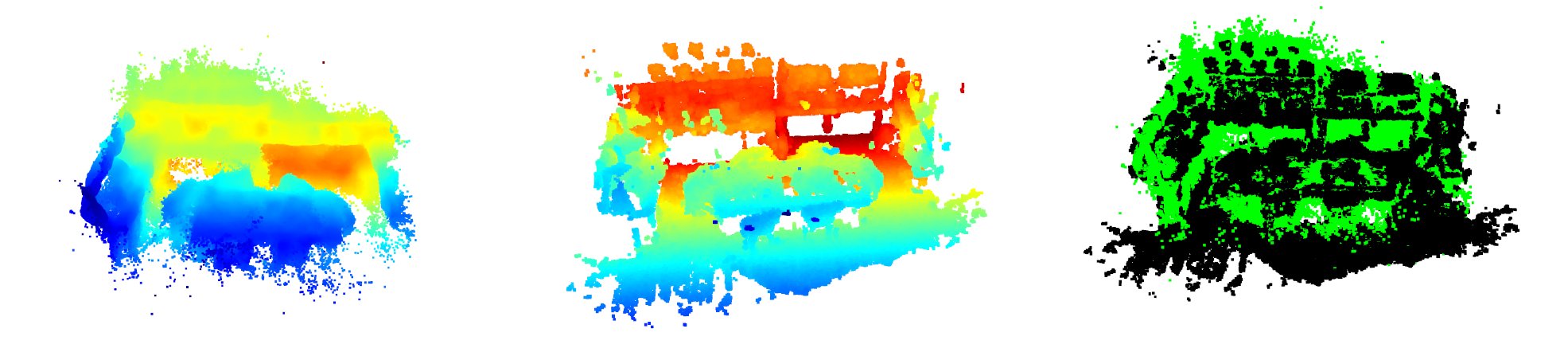}
    \caption{\textbf{Scene Geometry.} We plot the 3D geometry of the redkitchen sequence of the 7-scenes dataset~\cite{DBLP:conf/cvpr/ShottonGZICF13}: \textit{(left)} point cloud generated by the network after training, \textit{(middle)} ground-truth point cloud, \textit{(right)} ground-truth point cloud (black) overlaid on learned point cloud (green) in the same coordinate system. The fact that the learned geometry converged to the ground-truth geometry strongly suggests the existence of implicit triangulation among similar descriptors. }
    \label{fig:ps_rq1:scenes}
\end{figure*}

Given a dataset of $N$ grayscale images $\mathcal{D} = \{\image_1, \image_2, \ldots, \image_N\}$ and an associated ground-truth map produced via a Structure-from-Motion (SfM) algorithm~\cite{wu2013towards, schoenberger2016sfm}, the problem is to estimate the 6-DoF camera pose $\extrinsic_i \in \text{SE}(3)$ for a given image $\image_i \in \mathcal{D}$. Here, the camera pose $\extrinsic_i$ is an element of \(\text{SE}(3)\) which refers to the special Euclidean group that represents rigid body transformations in 3D space, including both rotation and translation.

To accomplish this, we approximate a function $\Tilde{f}_{(\image, W)}=~X$, where $X \in \mathbb{R}^{3\times \frac{w}{8}\times \frac{h}{8}}$ contains the scene coordinates $\mathbf{x}_{ij} \in \mathbb{R}^3$ for every $j$-th pixel in downsampled image $\image_i$, using a neural network parameterized by weights $W$. Therefore, $f$ implements a mapping from grayscale images to 3D coordinates, $\Tilde{f}_{(\image, W)}: \mathbb{R}^{w\times h} \rightarrow \mathbb{R}^{3\times \frac{w}{8}\times \frac{h}{8}}$. The network is trained on images $\image_i \in \mathcal{D}$ and the corresponding ground-truth poses $\extrinsic_i$ by minimizing a re-projection objective given by:

\begin{equation}
\label{eq:first obj}
    r(\crd_{ij}, \extrinsic_i) = ||\hat{\mathbf{y}}_{ij} - K_i\extrinsic^{-1}_i\hat{\crd}_{ij}||_1,
\end{equation}
where $K_i \in \mathbb{R}^{3\times 3}$ is the  camera calibration matrix, $\pos_{ij} \in \mathbb{R}^{2}$ denotes the $j$-th pixel coordinate in the pixel grid of $\image_i$ and  $\crd_{ij} \in \mathbb{R}^{3}$ represents the $j$-th predicted scene coordinate. The hat operator $\hat{\cdot}$ denotes the homogeneous representation and $||\cdot||_1$ the L1-norm. While the fundamental objective remains consistent with Eq.~\ref{eq:first obj}, more complex objectives are used in practice; we refer the reader to~\cite{DBLP:journals/pami/BrachmannR22, brachmann2023accelerated} for more details.

Finally, the camera pose can be computed given pairs of 2D-3D correspondences $\{(\pixel_{ij},\crd_{ij})\}$ using an off-the-shelf Perspective-n-Point solver~\cite{PoseLib, persson2018lambda}.

%% file: 4-prelimilary.tex
\subsection{Accelerated Coordinate Encoding}
While FocusTune is a generic method that is in principle compatible with a range of SCR methods, we have chosen to demonstrate our sampling technique within the Accelerated Coordinate Encoding (ACE)~\cite{brachmann2023accelerated} framework that we briefly describe in this section. ACE partitions the neural network function $\Tilde{f}_{(\image, W)}$ into a convolutional backbone $f_{\text{B}}$ (which only needs to be trained once and remains the same across all scenes) and a multi-layer perceptron (MLP) head $f_{\text{H}}$ (which is scene specific). Formally, the convolutional backbone $f_{\text{B}}: \mathbb{R}^{h \times w} \rightarrow \mathbb{R}^{512 \times \frac{h}{8} \times \frac{w}{8}}$, pre-trained on ScanNet~\cite{dai2017scannet}, generates dense local feature descriptors $\desc_{ij} \in \mathbb{R}^{512}$ for a query image $\image_i$. Subsequently, each descriptor $\desc_{ij}$ is passed to the regression head $f_{\text{H}}: \mathbb{R}^{512 \times \frac{h}{8} \times \frac{w}{8}} \rightarrow \mathbb{R}^{3 \times \frac{h}{8} \times \frac{w}{8}}$ to generate scene coordinates $\mathbf{x}_{ij} \in \mathbb{R}^3$. Every scene coordinate $\mathbf{x}_{ij}$ corresponds to the pixel coordinate $\pixel_{ij} \in \mathbb{R}^2$ of the descriptor $\desc_{ij}$ to form a pair of 2D-3D correspondences $\{(\pixel_{ij},\crd_{ij})\}$ to re-localize the image $\image_i$.

For training, ACE employs a buffer $\buffer$ containing sampled feature descriptors along with their geometric constraints. Each instance $\instance_k \in \buffer$ is expressed as $(\desc_k, \pixel_k, \intrinsic_k, \extrinsic_k)$, where $\desc_k$ is the descriptor vector at pixel coordinate $\pixel_k$ in an image $\image$ with associated intrinsic matrix $\intrinsic_k$ and extrinsic matrix $\extrinsic_k$. Note that the index notation changes from $ij$ at the image level (Eq.~\ref{eq:first obj}) to $k$ at the buffer level because a fixed amount of pixels is sampled from a particular training image for the buffer.

For larger scenes like those in the Cambridge Landmarks dataset~\cite{DBLP:conf/iccv/KendallGC15}, it is beneficial to divide the scene into multiple smaller sub-scenes $m$ and train a separate $f_{\text{H}_m}$ for each of them. At run time, all $f_{\text{H}_m}$ models are queried and the prediction with the highest inlier count is selected~\cite{brachmann2023accelerated}.

%% file: 5-methodology.tex
\begin{figure*}[t]
    \centering
    \includegraphics[width=1\linewidth]{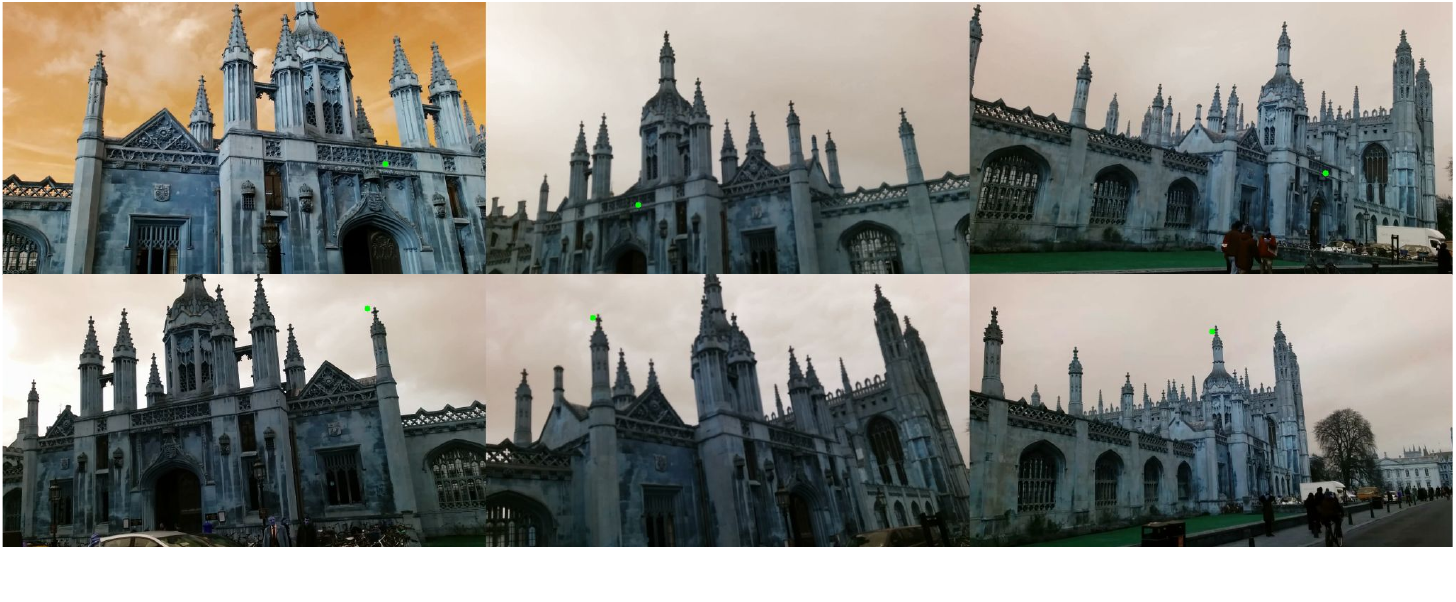}
    \vspace*{-1.25cm}
    \caption{\textbf{Examples of sampled instances with similar descriptors in the buffer $\buffer$.} The green dot denotes the projection of the network prediction given the descriptor. \textit{(Top row)} A unique appearance point leads to successful triangulation with an average re-projection error of 0.74 pixels. \textit{(Bottom row)} An ambiguous appearance point leads to unsuccessful triangulation with an average re-projection error of 72.3 pixels.}
    \label{fig:ps_rq1:appearance-example}
\end{figure*}

\section{Methodology}
\subsection{Implicit Triangulation}
The effectiveness of Accelerated Coordinate Encoding (ACE) is inherently linked to its ability to infer ground-truth scene geometry with precision (see Fig.~\ref{fig:ps_rq1:scenes}). While the re-projection error in Eq.~\ref{eq:first obj} does not explicitly address the depth attributes of the descriptors, the network function $f_{\text{H}}$ is conditioned to perform implicit triangulation through the training buffer $\buffer$. Specifically, $f_{\text{H}}$ is required to produce consistent scene coordinates for similar descriptors $(\desc_1, \desc_2, \desc_3, \dots)$ by leveraging the geometric constraints such as $\{(\pixel_1, \intrinsic_1, \extrinsic_1), (\pixel_2, \intrinsic_2, \extrinsic_2), (\pixel_3, \intrinsic_3, \extrinsic_3), \dots\}$.

Triangulation becomes feasible only when the 3D lines parameterized by the geometrical constraints $\{(\pixel_k, \intrinsic_k, \extrinsic_k)\}$ intersect at a common 3D point. However, tracking the feasibility of the geometrical constraints is intractable since there is no easy way to know exactly which constraints take part in the triangulation of a particular point.

We made a key observation that triangulation is often possible for 3D points with unique repeatable appearances rather than texture-less or ambiguous regions (refer to Fig.~\ref{fig:ps_rq1:appearance-example}). To identify such regions from the training images, a straightforward strategy might involve off-the-shelf feature detectors~\cite{detone2018superpoint, mishchuk2017working} or foreground/background segmentation models~\cite{lseg, kirillov2023segany}. However, during our experiments, these methods did not consistently yield effective results. On the other hand, the SfM program already employed robust local feature methods~\cite{DBLP:journals/ijcv/Lowe04, DBLP:conf/eccv/BayTG06, detone2018superpoint, mishchuk2017working, dusmanu2019d2} to identify these points (Fig.~\ref{fig:ps_rq1:appearance-example}, top row) during the reconstruction phase. Therefore, instead of employing a third-party system to identify reliable features for the training buffer $\buffer$, we rely on the SfM map to reduce the likelihood of sampling infeasible triangulation (Fig.~\ref{fig:ps_rq1:appearance-example}, bottom row).

\subsection{Guided Buffer Sampling}
To alleviate the issues associated with non-distinctive areas of the scene, we introduce a simple sampling strategy for populating the buffer $\buffer$. For each query image $\image_i$, we employ a pre-defined set of seed keypoints $\pixels_i$ that are generated by re-projecting the corresponding 3D points from the SfM map into the 2D image plane using the ground-truth camera poses (see Fig.~\ref{fig:ps_rq1:system_overview}).

To generate the set of seed keypoints $\pixels_i$ for each image $\image_i$, we initially retrieve the 3D coordinates $\crds_i$ of the 3D points that are visible in the image $\image_i$ from the SfM map. These coordinates are re-projected into the 2D image plane using the ground-truth intrinsic and extrinsic matrices, resulting in $\pixels_i = \intrinsic_i \extrinsic_i^{-1} \crds_i$.

\begin{figure*}[ht]
    \centering
    \includegraphics[width=\textwidth]{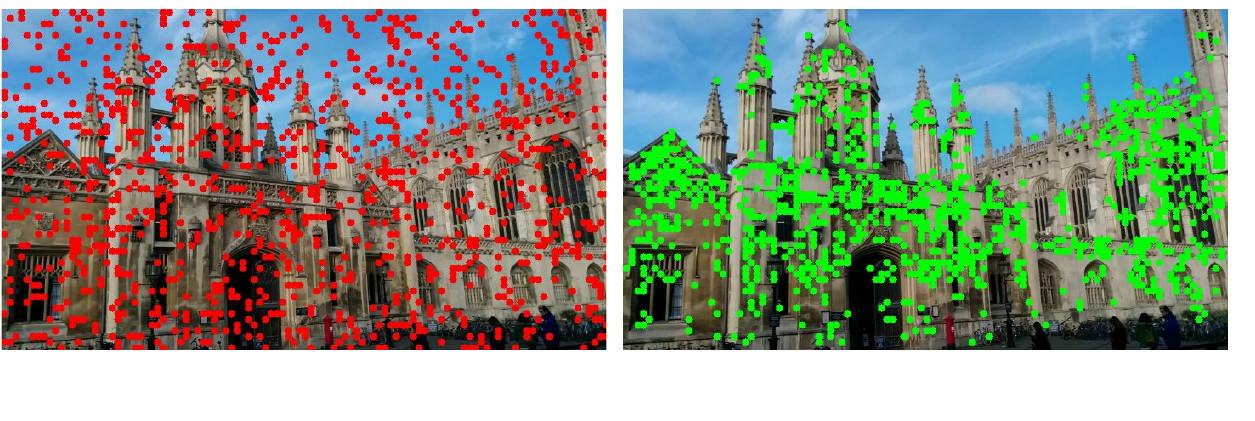}
    \vspace*{-1.5cm}
    \caption{\textbf{Sampling strategies comparison.} \textit{(Left image)} An example of the sampled pixels (red dots) using random sampling. \textit{(Right image)} An example of the sampled pixels (green dots) using our sampler. Our sampler focuses on important building structures instead of unimportant, texture-less regions (e.g., the sky) of the image. }
    \label{fig:sampler}
\end{figure*}

Due to random rotations and resizing applied to the camera matrices due to the data augmentation process during training time, some of the re-projected pixel coordinates may fall outside the image frame. Such invalid coordinates are then discarded, retaining only the valid set as the seed keypoints. Each seed keypoint forms a circle of radius $\radius$ pixels around it. Within these circles, keypoints are sampled uniformly, while areas outside the circles are excluded from sampling. This targeted sampling technique allows the network to concentrate on semantically relevant portions of the image, such as architectural features, rather than non-distinctive, texture-less zones (e.g., sky). We refer the reader to Fig.~\ref{fig:sampler} for a comparison of both sampling strategies.

%% file: 6-results.tex
\begin{table*}[t]
\centering
\footnotesize
\resizebox{\textwidth}{!}{%
\begin{tabular}{clcccccccccccc}
\toprule
\multicolumn{1}{l}{} & & & & & \multicolumn{8}{c}{7 Scenes} & \multicolumn{1}{l}{} \\
\cmidrule{6-13} 
\multicolumn{1}{l}{} & & \multirow{-2}{*}{\begin{tabular}[c]{@{}c@{}}{Mapping w/}\\ {Mesh/Depth}\end{tabular}} & \multirow{-2}{*}{\begin{tabular}[c]{@{}c@{}}{Mapping}\\ {Time}\end{tabular}} & \multirow{-2}{*}{\begin{tabular}[c]{@{}c@{}}{Map}\\ {Size}\end{tabular}} & {Pumpkin} & {Redkitchen} & {Stairs}  & {Office} & {Heads} & {Chess} & {Fire} & {Average} \\
\midrule
\multirow{3}{*}{\shortstack[c]{Feature\\Matching}} & AS (SIFT) \cite{active_search} &  No &   &  $\sim$200MB & 99.6\% & 99.8\% & 91.9\% & 98.6\% & 100\% & 99.9\% & 99.8\% & 98.5\% \\
 & D.VLAD+R2D2 \cite{kapture2020}  &  No &  $\sim$1.5h &  $\sim$1GB & 98.8\% & 98.4\% & 76.9\% & 99.7\% & 97.0\% & 100\% & 100\% & 95.7\% \\
 & hLoc (SP+SG) \cite{sarlin2019coarse, sarlin2020superglue} &  No &   &  $\sim$2GB & 100\% & 98.6\% & 72.0\% & 100\% & 100\% & 100 \% & 99.4\% & 95.7\% \\
\midrule
SCR & DSAC* \cite{DBLP:journals/pami/BrachmannR22} &  Yes &  15h &  28MB & 90.9\% & 96.4\% & 88.4\% & 95.3\% & 99.5\% & 99.6\% & 96.6\% & 95.3\%  \\

(w/Depth) & SRC \cite{dong2022visual} &  Yes &  2 min &  40MB & - & - & - & - & - & - & - & 81.1\%  \\
\midrule
& DSAC* \cite{DBLP:journals/pami/BrachmannR22} &  No &  15h &  28MB &  99.0\% & 97.0\% & \textbf{92.0\%} & 98.1\% & 99.8\% & 99.9\% & 98.9\% & {\ul 97.8\%}  \\
SCR & ACE \cite{brachmann2023accelerated} &  No &  5 min &  4MB & \textbf{99.9\%} & {\ul 98.5\%} & 83.1\% & \textbf{100\%} & \textbf{99.9\%} & \textbf{100\%} & {\ul 99.4\%} & 97.2\%  \\
& FocusTune (ours) &  No &  6 min &  4MB & {\ul 99.7\%} & \textbf{99.0\%} & {\ul 87.1\%} & {\ul 99.9\%} & \textbf{99.9\%} & \textbf{100\%} & \textbf{99.5\%} & \textbf{97.9\%}  \\
\bottomrule
\end{tabular}
}
\caption{\textbf{Indoor relocalization on 7-scenes~\cite{DBLP:conf/cvpr/ShottonGZICF13}.} We report the percentage of test frames below a 5 cm and 5 degrees pose error (higher is better). Best results are shown in \textbf{bold} for the Scene Coordinate Regression (``SCR'') group, second best results are \underline{underlined}. We improve upon the performance of ACE~\cite{brachmann2023accelerated}, especially in the difficult Stairs scene by an absolute 4.0\% improvement, and achieve slightly higher performance than DSAC*~\cite{DBLP:journals/pami/BrachmannR22} on average. The results for baseline methods were obtained from~\cite{brachmann2023accelerated,DBLP:conf/iccv/BrachmannHRS21}.}
\label{tab:results_indoor}
\vspace{-1em}
\end{table*}

\begin{table}[th]
  \centering
  \footnotesize
\begin{tabular}{lccccccccc}\toprule
& {ACE}~\cite{brachmann2023accelerated} & {FocusTune (ours)} \\
&$^\circ/$cm&$^\circ/$cm\\
           \midrule
Pumpkin    & 0.21 / 1.06 & \textbf{0.20 / 1.00}     \\
Redkitchen & 0.20 / 0.77 & \textbf{0.18 / 0.73}\\
Stairs     & 0.82 / 2.81 & \textbf{0.76 / 2.60}  \\
Office     & 0.28 / 1.04 & \textbf{0.24 / 0.97}  \\
Heads      & 0.33 / 0.53 & \textbf{0.29 / 0.50} \\
Chess      & 0.18 / 0.54 & \textbf{0.17 / 0.50} \\ 
Fire       & 0.32 / 0.82 & \textbf{0.30 / 0.77} \\
\midrule
Average    & 0.33 / 1.08 & \textbf{0.30 / 1.01}  \\

\bottomrule
\end{tabular}

  \caption{\textbf{7-scenes dataset~\cite{DBLP:conf/cvpr/ShottonGZICF13}}. We report the median rotation (in degrees) and translation error (in cm) for the ACE baseline method~\cite{brachmann2023accelerated} and our proposed FocusTune (lower is better). Best results are shown in \textbf{bold}. We consistently outperform ACE in all scenes, on average by 0.03 degrees and 0.07 cm.}
  \label{tab:results-7scenes}
\end{table}

\begin{table*}[t]
\centering
\footnotesize
\resizebox{\textwidth}{!}{%
\begin{tabular}{clccccccccc}
\toprule
\multicolumn{1}{l}{} & & & & & \multicolumn{5}{c}{Cambridge Landmarks} \\
\cmidrule(l){6-10} 
\multicolumn{1}{l}{} & & \multirow{-2}{*}{\begin{tabular}[c]{@{}c@{}}Mapping w/\\ Mesh/Depth\end{tabular}} & \multirow{-2}{*}{\begin{tabular}[c]{@{}c@{}}Mapping\\ Time\end{tabular}} & \multirow{-2}{*}{\begin{tabular}[c]{@{}c@{}}Map\\ Size\end{tabular}} & Court & King's & Hospital & Shop & St. Mary's 
& \multirow{-2}{*}{\begin{tabular}[c]{@{}c@{}}Average \\ (cm / $^\circ$)\end{tabular}} \\ 
\midrule
\multirow{5}{*}{\shortstack[c]{Feature\\Matching}} & AS (SIFT) \cite{active_search} &  No &   &  $\sim$200MB & 24/0.1 & 13/0.2 & 20/0.4 & 4/0.2 & 8/0.3 & 14/0.2 \\
 & hLoc (SP+SG) \cite{sarlin2019coarse, sarlin2020superglue} &  No &   &  $\sim$800MB & 16/0.1 & 12/0.2 & 15/0.3 & 4/0.2 & 7/0.2 & 11/0.2 \\
 & pixLoc \cite{sarlin21pixloc} &  No &   &  $\sim$600MB & 30/0.1 & 14/0.2 & 16/0.3 & 5/0.2 & 10/0.3 & 15/0.2\\
 & GoMatch \cite{zhou2022geometry} &  No &   &  $\sim$12MB & N/A & 25/0.6 & 283/8.1 & 48/4.8 & 335/9.9 & N/A \\
 & HybridSC \cite{compression2019cvpr} &  No & \multirow{-5}{*}{ $\sim$35min} &  $\sim$1MB & N/A & 81/0.6 & 75/1.0 & 19/0.5 & 50/0.5 & N/A \\
\midrule
\multirow{2}{*}{APR} & PoseNet17 \cite{kendall2017geometric} &  No &  4 -- 24h &  50MB & 683/3.5 & 88/1.0 & 320/3.3 & 88/3.8 & 157/3.3 & 267/3.0\\
 & MS-Transformer \cite{DBLP:conf/iccv/ShavitFK21} &  No &  $\sim$7h &  $\sim$18MB & N/A & 83/1.5 & 181/2.4 & 86/3.1 & 162/4.0 & N/A \\
\midrule
\multirow{3}{*}{\shortstack[c]{SCR\\w/Depth}}
 & DSAC* \cite{DBLP:journals/pami/BrachmannR22} &  Yes &  15h &  28MB & 49/0.3 & 15/0.3 & 21/0.4 & 5/0.3 & 13/0.4 & 21/0.3 \\
 & SANet \cite{yang2019sanet} &  Yes &  $\sim$1min &  $\sim$260MB & 328/2.0 & 32/0.5 & 32/0.5 & 10/0.5 & 16/0.6 & 84/0.8\\
& SRC \cite{dong2022visual} &  Yes &  2 min &  40MB & 81/0.5 & 39/0.7 & 38/0.5 & 19/1.0 & 31/1.0 & 42/0.7\\
\midrule
\midrule
\multirow{6}{*}{SCR}
 & DSAC* \cite{DBLP:journals/pami/BrachmannR22} &  No &  15h &  28MB & 34/0.2 & {\ul 18/0.3} & {\ul 21/0.4} & {\ul 5/0.3} & 15/0.6 & 19/0.4 \\
& ACE \cite{brachmann2023accelerated} &  No &  5 min &  4MB & {43/0.2} & 28/0.4 & {31/0.6} & {\ul 5/0.3} & {18/0.6} & 25/0.4\\
& FocusTune (ours) &  No &  6 min &  4MB & {38/0.1} & 19/0.3 & {18/0.4} & 6/0.3 & {15/0.5} & 19/0.3\\
\cmidrule{2-11}
& ACE (4 model ensemble) \cite{brachmann2023accelerated} &   No &  20 min &  16MB & \textbf{28/0.1} & {\ul 18/0.3} & 25/0.5 & {\ul 5/0.3} & \textbf{9/0.3} & {\ul 17/0.3} \\
& FocusTune (4 model ensemble; ours) &   No &  24 min &  16MB & {\ul 29/0.1} & \textbf{15/0.3} & \textbf{17/0.4} & \textbf{5/0.2} & \textbf{9/0.3} & \textbf{15/0.3} \\

\bottomrule
\end{tabular}
}
\caption{\textbf{Cambridge Landmarks \cite{DBLP:conf/iccv/KendallGC15} Results.} We report median rotation (in degrees) and position errors (in cm). Best results for the Scene Coordinate Regression (SCR) methods are shown in \textbf{bold}, and second best results are \underline{underlined}. Our proposed FocusTune outperforms the direct competitor ACE in all but one scenes, both for a single model and a 4 model ensemble. The 4 model ensemble of FocusTune also consistently outperforms  DSAC* which has a significantly higher mapping time, closing the gap to structured methods that employ feature matching. The results for baseline methods were obtained from~\cite{brachmann2023accelerated}.%
}
\label{tab:results_cam}
\end{table*}

\section{Experimental Setup and Results}
This section extensively evaluates the proposed FocusTune sampling method. We first provide details on the implementation in Section~\ref{subsec:impl}, followed by an overview of the evaluation datasets in Section~\ref{subsec:datasets}. This is followed by a rigorous validation of FocusTune's effectiveness using two diverse datasets: the 7-scenes dataset for small-scale indoor environments and the Cambridge Landmarks Dataset for large-scale outdoor settings (Section~\ref{subsec:sota}). We then assess the sensitivity of our method to variations in the sampling radius in Section~\ref{subsec:sensitivity}. Finally, Section~\ref{subsec:samplingstrategy} provides additional insights into the focus-guided sampling strategy. Overall, our findings demonstrate the significant localization accuracy of FocusTune over existing methods while retaining the computational efficiency of ACE models~\cite{brachmann2023accelerated}.

\subsection{Implementation}
\label{subsec:impl}
We incorporate our FocusTune sampling approach into the ACE learning pipeline as described in Brachmann et al.~\cite{brachmann2023accelerated}. Our training buffer consists of 8 million instances. Keeping consistent with ACE's original settings, we optimise $f_\text{H}$ over 16 complete iterations using a batch size of 5120. For comprehensive details on other hyper-parameters, the reader is referred to \cite{brachmann2023accelerated}. Unless otherwise mentioned, we set the sampling radius $\radius=5$ which is the only hyperparameter of our method.

\subsection{Datasets}
\label{subsec:datasets}
FocusTune requires ground-truth 3D models for training; therefore, our evaluations are based on two datasets that provide such ground-truth information:

The \textbf{7-scenes dataset}~\cite{DBLP:conf/cvpr/ShottonGZICF13} provides ground-truth camera poses for small-scale indoor scenes using a Kinect RGB-D camera with $640\times 480$ resolution. Even though the depth information is provided for each image, we only use RGB frames to train $f_\text{H}$. Seven sequences were captured for each scene by different users and then were split into separate sets for training and testing purposes. Ambiguities, such as repeated steps in the Stairs scene, as well as specularities like reflective cupboards in the RedKitchen scene, are present in both RGB and depth images. Other challenges include motion blur, different lighting conditions, flat surfaces, and sensor noise. The ground-truth camera poses and 3D point cloud for each sequence were obtained using COLMAP~\cite{schoenberger2016sfm} and provided by~\cite{DBLP:conf/iccv/BrachmannHRS21}. With the sampling radius $\radius=5$, our sampling method makes use of approx. $15.4\%$ of the pixels in the training images on average. 

The \textbf{Cambridge Landmarks Dataset}~\cite{DBLP:conf/iccv/KendallGC15} was collected at 5 different sites at the University of Cambridge in a large-scale outdoor urban setting with a Google LG Nexus 5 smartphone into high-definition videos. These videos were subsampled at a 2 Hz rate to obtain RGB frames. Substantial urban obstacles like pedestrians and vehicles were evident, with data being gathered across various timeframes that reflected diverse lighting and weather situations. Training and testing sets came from separate walking trajectories rather than being selected from a single trajectory, thereby rendering it more challenging for the localization task. The ground-truth training and testing camera poses as well as the 3D models were reconstructed using VisualSfM~\cite{wu2013towards} and made available with the dataset. With sampling radius $\radius=5$, our sampling method makes use of approximately $16\%$ of the pixels in the training images on average. 

\subsection{Comparison to State-of-the-Art}
\label{subsec:sota}
\subsubsection{7-scenes Dataset} We now compare FocusTune to the state-of-the-art on the 7-scenes indoor dataset. Employing FocusTune resulted in models that consistently outperform the ACE baseline~\cite{brachmann2023accelerated} across all tested indoor scenes, highlighted by either higher percentages of test images localized within 5$^\circ$ and 5 cm of the ground truth poses (Table~\ref{tab:results_indoor}) or lower median errors when the percentages are comparable (Table~\ref{tab:results-7scenes}). Notably, in texture-deficient scenes such as the Stairs sequence, we observed an improvement of approximately $4\%$ over ACE. However, for easier scenes, our sampling method also leads to even more accurate localizers with lower median errors across all test sequences (Table~\ref{tab:results-7scenes}).

\subsubsection{Cambridge Landmarks Dataset} As shown in Table~\ref{tab:results_cam}, compared to ACE~\cite{brachmann2023accelerated}, FocusTune excelled on the Cambridge landmarks dataset with a significant median error reduction of $0.1^\circ$ in rotation and $6$ cm in translation for a single model. The improvement was even more pronounced in challenging scenes like the King (translation error reduction of 9~cm) and Hospital (translation error reduction of 13~cm) sequences.

Similar observations hold for the 4 model ensemble variants, where FocusTune improved upon ACE's performance by on average 2~cm. In scenes like the King and Hospital sequences, our method again reduces the translation errors from 18~cm to 15~cm and 25~cm to 17~cm, respectively. FocusTune thus closes the gap further to structured methods relying on feature matching~\cite{sarlin2019coarse, sarlin2020superglue,active_search,sarlin21pixloc}, while retaining the significantly low storage requirements of ACE~\cite{brachmann2023accelerated}, and performs better than DSAC*~\cite{DBLP:journals/pami/BrachmannR22}.

\subsection{Sampling Radius}
\label{subsec:sensitivity}
\begin{figure}[t]
    \centering
    \includegraphics[width=0.98\linewidth]{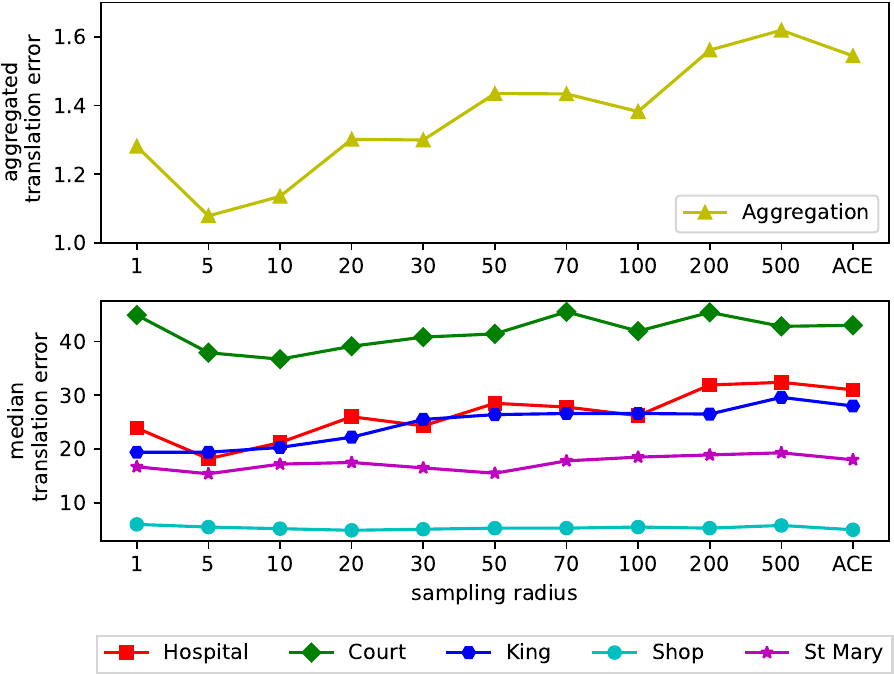}
    \caption{\textbf{Ablation studies on $\radius$.} We plot the normalized aggregated (top) and absolute individual (bottom) median translation errors for our method, using various sampling radii $\radius$, across the five sequences of the Cambridge Landmarks dataset~\cite{DBLP:conf/iccv/KendallGC15}. A sampling radius of $\radius=5$ leads to highest performance, with smaller values degrading in performance as the number of keypoints reduces too much. Larger sampling radii include areas that are not informative, and similarly result in reduced performance.}
    \label{fig:ablation}
\end{figure}

In this section, we assess the sensitivity of our method to variations in the sampling radius $\radius$. The bottom plot in Fig.~\ref{fig:ablation} illustrates the median translation error for the sequences from the Cambridge Landmarks dataset \cite{DBLP:conf/iccv/KendallGC15}. 

To determine the operational radius for our experiments, we aggregate the results from the five sequences using the following approach: we individually normalize the results of each sequence with its corresponding minimum error. For each radius value, we calculate the mean of the normalized errors across the different sequences and add one standard deviation. The upper plot in Fig.~\ref{fig:ablation} presents the resulting normalized error, which combines the outcomes from all five sequences.

For the remainder of our experiments, we select a sampling radius of $\radius = 5$, which corresponds to the point where our approach demonstrates optimal performance with the lowest error. We observe that larger values result in decreased performance because the buffer becomes populated with keypoints from non-textured areas. Note that $ \radius \to \infty$ falls back to the default ACE random sampling. Conversely, smaller values lead to reduced performance due to a decrease in the number of keypoints.

\subsection{Sampling Strategy Analysis}
\label{subsec:samplingstrategy}
\begin{figure}[t]
    \centering
    \includegraphics[width=1.0\linewidth]{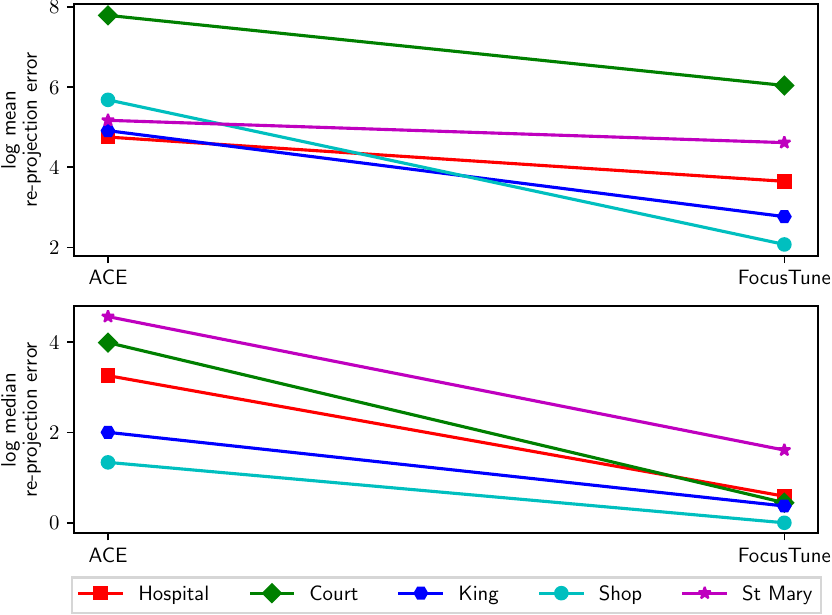}
    \caption{\textbf{Training efficiency.} We sample two buffers from the Cambridge Landmarks dataset~\cite{DBLP:conf/iccv/KendallGC15}: $\buffer_{\text{ACE}}$ using random sampling as in ACE~\cite{brachmann2023accelerated} (\textit{left}) and $\buffer_{\text{FocusTune}}$ using our proposed focus-guided sampling strategy (\textit{right}). We compute the mean and median re-projection errors of all the instances after training and plot the log of the errors. The decrease in training re-projection errors suggests that $f_\text{H}$ is able to triangulate more points from $\buffer_{\text{FocusTune}}$ than $\buffer_{\text{ACE}}$, highlighting a better use of the dataset.}
    \label{fig:efficiency}
\end{figure}

To elucidate the efficiency of our sampling method, we compared two training buffers: $\buffer_\text{ACE}$ using the random sampling strategy employed by ACE and $\buffer_\text{FocusTune}$ using our sampling strategy. Both mean and median re-projection errors dropped markedly across all scenes when employing our sampling strategy, suggesting more effective dataset utilization (Fig.~\ref{fig:efficiency}).

%% file: 7-conclusion.tex
\section{Conclusions and Future Works}
In this paper, we introduced FocusTune, a novel focus-guided sampling strategy designed to improve the performance of visual localization algorithms. By leveraging key geometric constraints to guide the scene coordinate regressor, FocusTune advances the state-of-the-art in visual localization accuracy while reducing computational and storage requirements. Notably, we demonstrated these benefits by integrating FocusTune into the recently developed Accelerated Coordinate Encoding (ACE) model, achieving reductions in translation errors on indoor and outdoor benchmark datasets.

Even though SfM maps often come with ground-truth camera poses in common capturing processes~\cite{schoenberger2016sfm, wu2013towards} and SfM models are highly useful for downstream applications (e.g., augmented reality), access requirements for these models might render our method inapplicable for scenarios in which the mapping poses are captured via depth SLAM~\cite{dai2017bundlefusion}. However, our experiments show that SCR models benefit from training with salient image regions that can still be discovered using an off-the-shelf feature detector\cite{detone2018superpoint, DBLP:journals/ijcv/Lowe04, DBLP:conf/eccv/BayTG06, mishchuk2017working, dusmanu2019d2}.

We suggest a range of promising directions to investigate for future works. Exploring adaptive sampling techniques could provide an increased level of adaptability to varying scene complexities. Whilst FocusTune is highly performant in itself, by integrating FocusTune's heuristic sampling strategy with other localization methods it may be possible to obtain further performance improvements, where this is practically acceptable. Finally, the exploration of geometrically-informed loss functions could further optimize the triangulation process, leading to more robust performance across a wider range of scenarios.

We believe that FocusTune represents a significant advancement in the field, offering both high performance and efficiency, thus making it particularly appealing for applications in mobile robotics and augmented reality.